\documentclass{article}

\usepackage{arxiv}

\usepackage[utf8]{inputenc} 
\usepackage[T1]{fontenc}    
\usepackage{hyperref}       
\usepackage{url}            
\usepackage{booktabs}       
\usepackage{amsfonts}       
\usepackage{nicefrac}       
\usepackage{microtype}      
\usepackage{lipsum}
\usepackage{amsmath}
\usepackage{graphicx}
\graphicspath{ {./images/} }
\usepackage{float}
\usepackage{verbatim}
\usepackage{multirow}
\usepackage{arydshln}
\usepackage{authblk}
\usepackage{enumitem}

\newcommand{\email}[1]{\href{mailto:#1}{\tt{\nolinkurl{#1}}}}
\newcommand{\orcid}[1]{ORCID: \href{https://orcid.org/#1}{\tt{\nolinkurl{#1}}}}

\title{condLSTM-Q: A novel deep learning model for predicting Covid-19 mortality in fine geographical Scale}

\author[1,$\dagger$]{HyeongChan Jo}
\author[2,$\dagger$]{Juhyun Kim}
\author[3]{Tzu-Chen Huang}
\author[1,*]{Yu-Li Ni, M.D.}
\affil[1]{Division of Biology and Biological Engineering, Caltech}
\affil[2]{The Division of Physics,
Mathematics and Astronomy, Caltech}
\affil[3]{Walter Burke Institute for Theoretical Physics, Caltech}
\affil[*]{Corresponding author: \email{ynni@caltech.edu}}
\affil[$\dagger$]{\fontsize{10}{10}\selectfont These authors contributed equally to this work.}

\begin{document}
\maketitle

\begin{abstract}
Predictive models with a focus on different spatial-temporal scales benefit governments and healthcare systems to combat the COVID-19 pandemic. Here we present the conditional Long Short-Term Memory networks with Quantile output (condLSTM-Q), a well-performing model for making quantile predictions on COVID-19 death tolls at the county level with a two-week forecast window. This fine geographical scale is a rare but useful feature in publicly available predictive models, which would especially benefit state-level officials to coordinate resources within the state. The quantile predictions from condLSTM-Q inform people about the distribution of the predicted death tolls, allowing better evaluation of possible trajectories of the severity. Given the scalability and generalizability of neural network models, this model could incorporate additional data sources with ease, and could be further developed to generate other useful predictions such as new cases or hospitalizations intuitively.
\end{abstract}




\section{Introduction}

The coronavirus disease 2019 (COVID-19) pandemic has taken more than 1.1 million lives worldwide and more than 224,000 lives in the United States as of October 2020. Predicting the trend of the pandemic in terms of deaths and positive cases precisely has been crucial, as it allows the governments and healthcare systems to effectively distribute and prioritize resource allocations \cite{hospital_impact}.

Here we present the conditional Long Short-Term Memory networks with Quantile output (condLSTM-Q) model, a novel deep learning model predicting the spatial and temporal distribution of COVID-19 outbreak. The condLSTM-Q was developed during the Caltech COVID-19 Initiative \cite{caltech}, a campaign aimed to develop and explore novel models to complement the classic epidemiological models. The optimization goal for participating models was to predict daily death tolls attributed to COVID-19 in each county across the United States with a two-week future prediction window. In addition, the models were required to provide estimations of 10-quantiles (0.1 to 0.9 quantile with intervals of 0.1) as the outputs, as quantile prediction is useful when forecasting the extremes. \cite{quantile_heat, quantile_eco}. The county-level spatial resolution was selected to inform state officials and local governments so that they can make timely decisions on resource allocations and strengthen their pandemic preparedness. The condLSTM-Q was among one of the best performers by the end of the campaign in early June. To test the robustness of this model, the model was then deposited without architectural modifications since mid-June, 2020, and has continued to be trained and output predictions using updated data. Since its deposit, the model's predictions were comparable to the well-known, publicly available predictions by the Institute for Health Metrics and Evaluation (IHME) \cite{IHME} throughout May-October when this manuscript was written, showcasing the robustness of condLSTM-Q.


The foundation of the condLSTM-Q was based on the Long Short-Term Memory networks (LSTM), a common neural network model well-suited for time series predictions \cite{hochreiter_long_1997}. There are previous works of using LSTM and other neural networks for COVID-19 forecasting, such as risk assessment of countries \cite{pal_neural_2020}, and predicting national COVID-19 mortality rates \cite{melin_multiple_2020}. Nevertheless, to our knowledge no county-level, quantile prediction model existed when our prototype was developed. Lack of county-based predictions could hugely impact the prevention and control of COVID-19 pandemic at local governments, which plays pivotal roles in the pandemic preparedness in the United States. Making quantile predictions rather than single-point predictions also gives a sense of the distribution of the predicted values, which can help governments to understand the situation better. Thus we decided to explore if the LSTM-based model could provide good quantile predictions at the county level. We improved the classical LSTM model by adapting and building on a "conditional" LSTM architecture which takes in static data more naturally along with time series data \cite{condRNN, FeiFeiInspired}. In addition, we utilized the flexibility of neural networks to output a distribution of 10-quantiles that was required by the initiative.

Given the robust performance of condLSTM-Q and the fact that we have not exploited all possible variations of the model, this architecture from our pilot experiment is worthy of further exploration. Interdisciplinary collaboration between experts in machine learning and epidemiology would greatly facilitate building better variants that will provide both longer prediction windows and clearer interpretability.

\section{Methods}

\subsection{Data Sources and Data Preprocessing}
Our model includes a variety of data that is expected to have correlations with the death toll. The original source of each data is given in Section \ref{sec:materials}. The data used in our model fall into one of the following seven categories: (1) COVID-19 mortality and confirmed cases provided by New York Times; (2) demographics and local health resources such as age composition, mortality rate by diseases, and the number of hospitals, provided by the Yu group at University of California, Berkeley; (3) county-wise gross domestic product (GDP) from the Bureau of Economic Analysis; (4) population density and geographical data from 2010 Census; (5) mobility changes in response to COVID-19 provided by Descartes Labs; (6) policy actions in response to COVID-19 such as state of emergency declaration, safe-at-home order, and business closure, from Covidvis team at University of California, Berkeley \cite{covidvis} and the U.S. Department of Health and Human Services; (7) and the U.S. pneumonia and influenza mortality report from the National Center for Health Statistics Mortality Surveillance System in Centers for Disease Control and Prevention (CDC). All the data were county-level, and the latest start date of the time series data was March 1, 2020.

Preprocessing specifics for each dataset are listed in the following:
\begin{itemize}[leftmargin=.2in]
    \item Mobility data from the Descartes Labs was provided for 2,721 counties out of 3,114 counties in total. The missing 393 counties' mobility data were filled in with their corresponding state-level data. Missing dates in mobility data of Descartes Labs were interpolated with the data on the closest existing date of the same day in a week to reflect the weekly pattern inherent in the data. As of November 8, 2020, 14,049 values were missing, out of total 688,413 data points (about 2\%). They were filled in using either the aforementioned same-day interpolation or spline interpolation, and the data from the first and the last day of recording was repeated to fill the missing data before and after the existing data, if necessary. Spline interpolation has an advantage over the same-day interpolation in handling a large number of missing data over a long period of time, whereas the latter can capture the weekly pattern that cannot be maintained in the former method.
    \item Seasonality feature was extracted from the U.S. pneumonia and influenza data, under the premise that the COVID-19 will follow a seasonality of virus that becomes more dormant during the summer and severe in the winter. More specifically, the multiplicative seasonality was extracted from the state-wide pneumonia and influenza mortality rate during flu seasons from 2013 to 2020 provided by CDC \cite{P&Irate}.
    \item Among 64 features of the demographics and local health data, 43 features including the population estimate and the mortality rate of various underlying diseases were selected as the static features.
    \item For the policy actions features, declaration of the state of emergency, safe-at-home action, and the closure of inessential business were selected as the static features.
\end{itemize}
All numerical values were standardized before the model training step.


\subsection{Implementation of condLSTM-Q}

\begin{figure}[ht]
    \centering
    \includegraphics[width=0.75\textwidth]{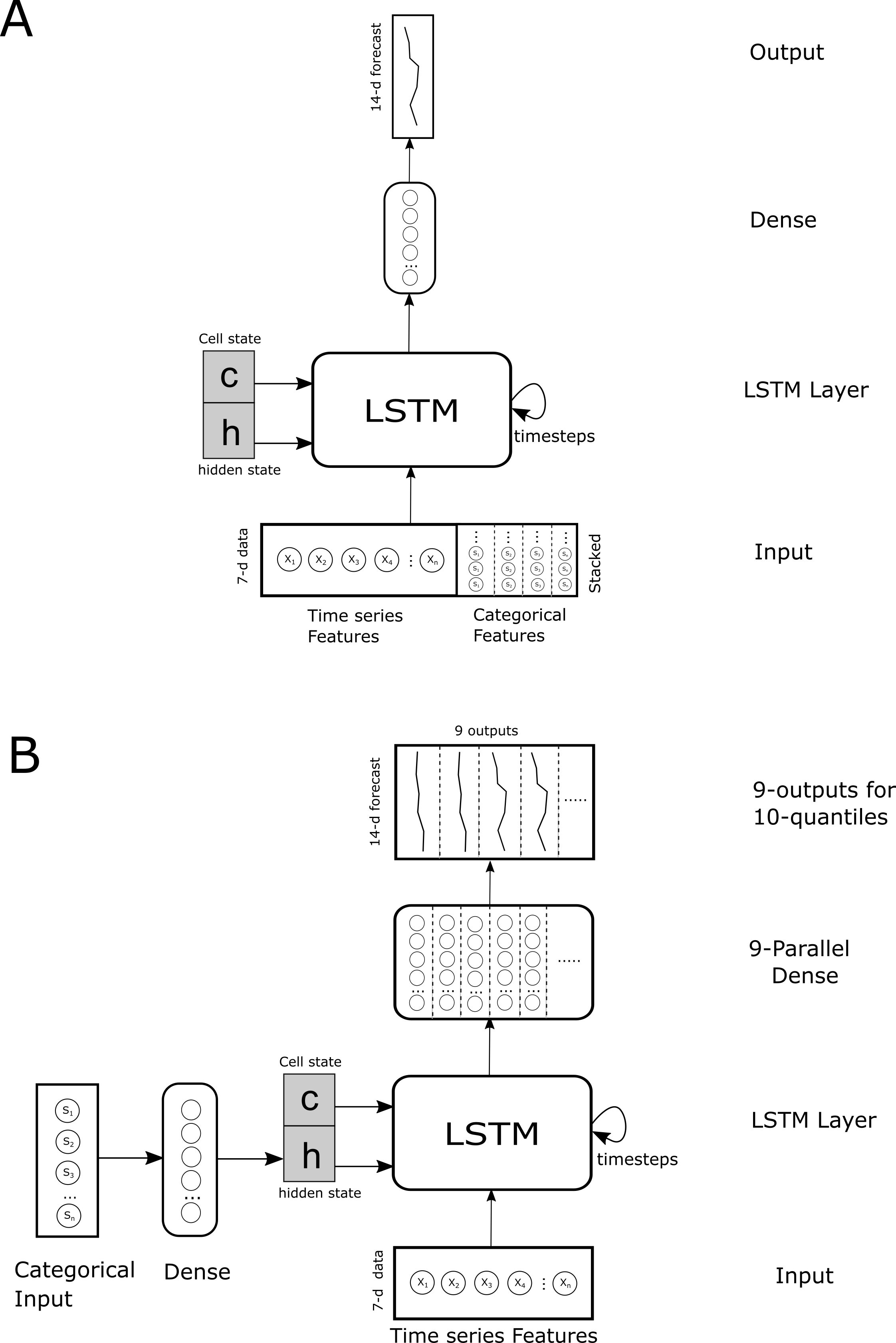}
    \caption{\textbf{Classical LSTM vs condLSTM-Q Architecture}. Classical LSTM architecture (Top). The classical LSTM can predict time series better than traditional RNN by remembering and passing cell states and hidden states over the timesteps. The hidden states are usually initialized to zero or random values. Categorical features have to be stacked and replicated to the same dimension to match the time series features. The classical structure outputs one time series. The condLSTM-Q architecture (Bottom). In condLSTM-Q, categorical inputs are passed through a fully connected layer as hidden states to the initial step ("Conditional" based on the static information) of the LSTM layer. A dropout layer is also applied to the dense layer to prevent overfitting to the categorical inputs. The output from the LSTM layer is fed into 9 parallel dense layers to provide 9 outputs for 10-quantiles ("Q"). Each of the 9 parallel outputs is given to the pinball loss function during training, thus the parallel dense layer gets updated to match the corresponding quantiles by minimizing the summed loss.}
    \label{fig:RNNarch}
\end{figure}

The classical LSTM which the condLSTM-Q's backbone was based on is shown in Fig \ref{fig:RNNarch}A. LSTM is a type of Recurrent Neural Network (RNN) that can be trained to learn the mapping from the time series features to the time series of interest. Compared to previous architectures of RNN, LSTM has the advantage of \textit{remembering} long-term dependencies by passing the information through its \textit{cell state} and \textit{hidden states} recurrently. However, the classical LSTM architecture cannot take into account non-time series data (hereinafter referred to as \textit{categorical} data) in a natural way; such features have to be stacked into the same dimension of the time series data during data processing as shown in Fig \ref{fig:RNNarch} A, in order to take both types of data into the input stream. This way of data processing undermines the optimal performance of LSTM.

One approach to overcome this limitation is to initialize the initial states of the LSTM units in response to the categorical data. Intuitively, the categorical data should provide a priori information to the prediction, rather than real-time information. Thus, a "conditional" module can be added to get these categorical data and feed in "priors" as the hidden states, as implemented in \cite{condRNN}, in contrast to the usual LSTM where the hidden states are initialized to zeros or random noise \cite{LSTM_noisyInitial}. 

This is the way we treat the categorical data in our model, \textit{conditional LSTM with Quantile output model} (condLSTM-Q; Fig \ref{fig:RNNarch} B). In this model, categorical inputs are passed through a fully connected layer into the model as the \textit{hidden states} to the initial step of the LSTM layer. Through this design, we could input 50 categorical data such as \textit{Income}, \textit{Age and Gender Distribution}, \textit{Medicare coverage}, and \textit{Population} in their original forms of one scalar per feature into our model. After the initialization step, the cell states and hidden states in LSTM were then updated as in the classical LSTM, based on 8 time series features including \textit{mobility of the population}, \textit{new cases per day}, and \textit{new deaths per day}.

Another feature of condLSTM-Q is that it generates predictions on 10-quantiles (``Q"). To get the sense of the distribution of the predicted values, which help us understand the situation and the quality of the prediction better (a forecast with higher variance results in huge uncertainty in whatever derivatives from the forecast), our model forecasts the death counts in each quantile from 0.1 to 0.9 with 0.1 increments by a variant of the multi-output LSTM. In this model, the outputs from the LSTM layer are fed into 9 parallel dense layers to generate a vector of dimension 9 for each day in the prediction, where each component of the output vector is a forecast of each quantile. A standard measure for the accuracy of such a quantile forecasting is the \textit{pinball loss} defined as
\begin{align}
    loss = \max(q \cdot \epsilon, (q -1)\cdot \epsilon),
\end{align}
where $0<q<1$ is the quantile and $\epsilon=y-\hat{y}$ is the difference between the true target value $y$ and the predicted target value $\hat{y}$. As such, our model is trained to minimize the average pinball loss
\begin{align}\label{quantileloss}
    \frac{1}{9}\sum_{i=1}^9 \mathrm{max}(q_i\cdot \boldsymbol{\epsilon_i}, (q_i-1)\cdot \boldsymbol{\epsilon_i}),
\end{align}
where $q_i=i/10, 1\leq i \leq 9$ and $\boldsymbol{\epsilon_i} = \boldsymbol{y}-\boldsymbol{\hat{y_i}}$ is the difference between the true target vector and the predicted target vector for quantile $q_i$. 

\subsection{Hyperparameters and the Training}

The time series data after preprocessing is an array of shape $(\texttt{\#} \textrm{counties}, \texttt{\#}\textrm{dates}, \texttt{\#} \textrm{features})$, and the categorical data is an array of shape $(\texttt{\#} \textrm{counties}, \texttt{\#} \textrm{features})$. For each date in the time series data, the data is further split into a history window of size $7$ and a target window of size $14$, so that condLSTM-Q can learn to predict the mortality rate for the next 14 days based on the history of prior 7 days.

For hyperparameter tuning, we held out the last $21$ days' worth of data for validation. As of August 4, 2020, we used 136 days' worth of data until July 14, 2020, for the training set, and the remaining 21 days' worth of data for the validation set. Since the data was from 3,114 counties, and the input and the target for the model should be 7 and 14 days long, this led to 316,224 samples for training and 3,114 samples for validation.

The optimal hyperparameters selected were 128 units in LSTM, a learning rate of 0.001, and a dropout rate of 0.2. In this setting, overfitting was observed after $~30$ epochs, so the model was trained with the ADAM optimizer \cite{adam} for 20 epochs to avoid overfitting.

\section{Result}
\label{sec:headings}

\subsection{Predictions and Performance}

\begin{figure}[h]
    \centering
    \includegraphics[width=0.95\textwidth]{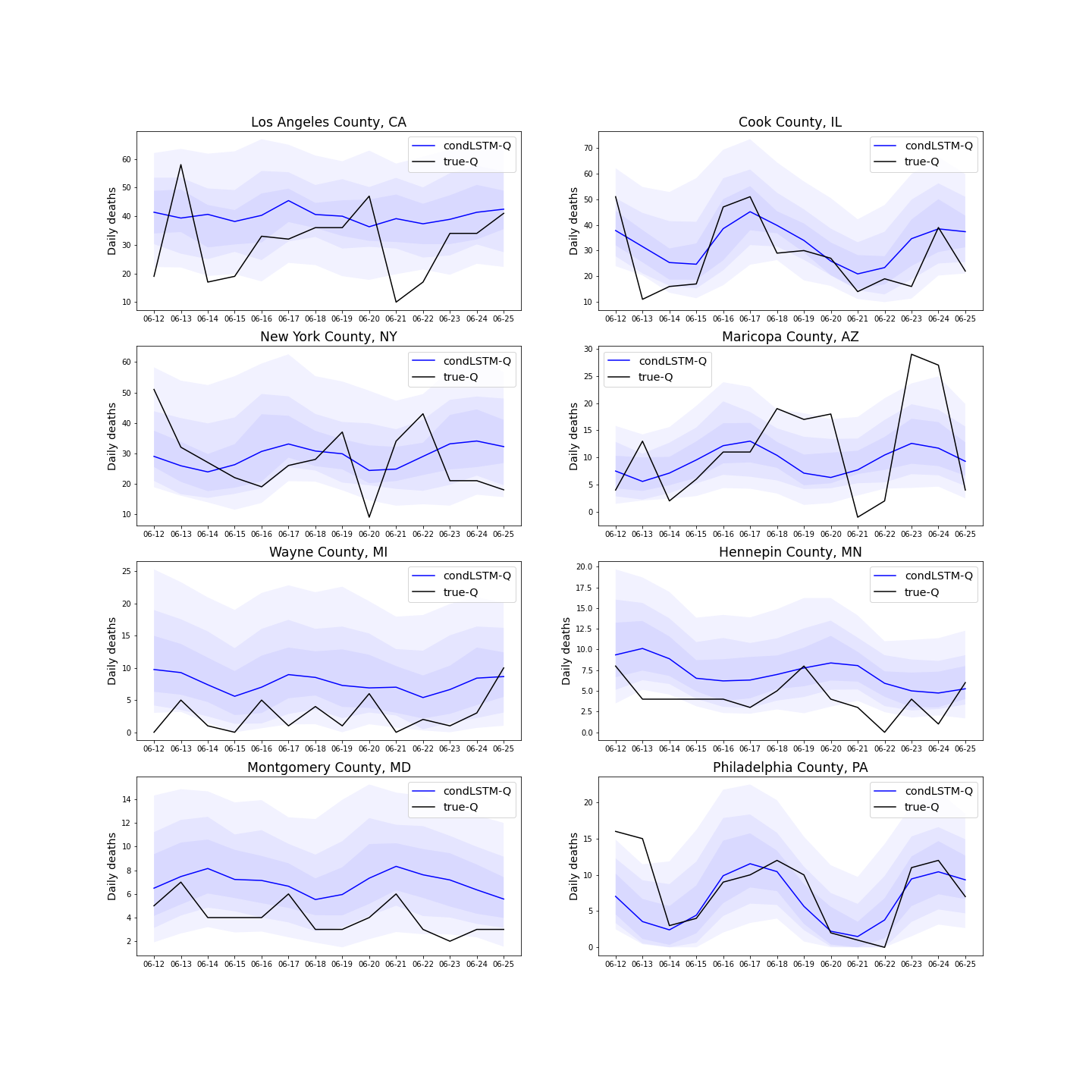}
    \caption{\textbf{Predictions in representative locations} Representative counties including Cook, Los Angeles, New York, Wayne, Philadelphia, Hennepin, Maricopa, and Montgomery. The condLSTM-Q was able to keep track of the death trends in different phases in the pandemic with different dynamic ranges. For instance, Los Angeles had daily deaths fluctuating between 20 - 60 per day whereas Philadelphia counts were roughly a third of that.}
    \label{fig:condFips}
\end{figure}

The condLSTM-Q provides a 14-day, county-level prediction with 10-quantiles. To illustrate this, we show predictions for several representative counties, identified by their Federal Information Processing Standards- or FIPS-based county code, including Cook County, Los Angeles, New York, Wayne, Philadelphia, Hennepin, Maricopa and Montgomery from June 12 to June 25, 2020 (See Fig \ref{fig:condFips}). This prediction interval was immediately after we finalized and froze the model architecture. The model was trained with data up to June 11, 2020, and was agnostic to the data afterward (i.e. post-June 12). The condLSTM-Q was able to predict the death trends in different phases in the pandemic with different dynamic ranges. For example, the number of daily deaths in Los Angeles fluctuated between 20 and 60 per day whereas the counts in Philadelphia were roughly a third of that. 

\begin{figure}
\centering
 \includegraphics[width=.8\textwidth]{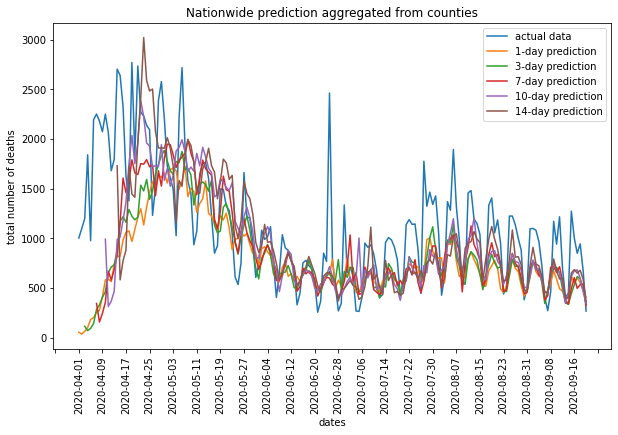}
   \caption{\textbf{Nationwide prediction on the total number of deaths, aggregated from counties.} At each time point, the actual data from that day is shown in blue, and the predicted values returned from models trained until 1, 3, 7, 10, and 14 days ago are shown in different colors. The model's overall accuracy is relatively low in the beginning when there was not enough data for the model to be trained on, but the prediction started to follow the trend well since late April}
   \label{fig:National}
\end{figure}

To validate the performance of the condLSTM-Q, we aggregated the sum of reported mortality cases of the counties and plotted the national trend with New York Times' statistics (Fig \ref{fig:National}). At first, the precision of predictive values was subjected to limited data for model training. The predictions gradually improved around mid-April, and successfully predicted the descending trend of COVID-19 mortality from May to July 2020 and the ``second wave” of COVID-19 since late July 2020. To observe the changes in prediction accuracy from day 1 to day 14 of the prediction (i.e. across the two-week prediction window), we aligned the day 1, 3, 7, 10, and 14 of each of the prediction windows to their corresponding dates. The overall prediction accuracy was improved with more available training data. Before mid-April 2020, distal prediction (e.g. day 14) was much more varied compared to proximal predictions; however, after the model became stable with enough training data, the performance of both distal and proximal prediction converged with a smaller spread, and the precision of distal prediction was not inferior to proximal predictions.

\begin{figure}
\centering
 \includegraphics[width=.6\textwidth]{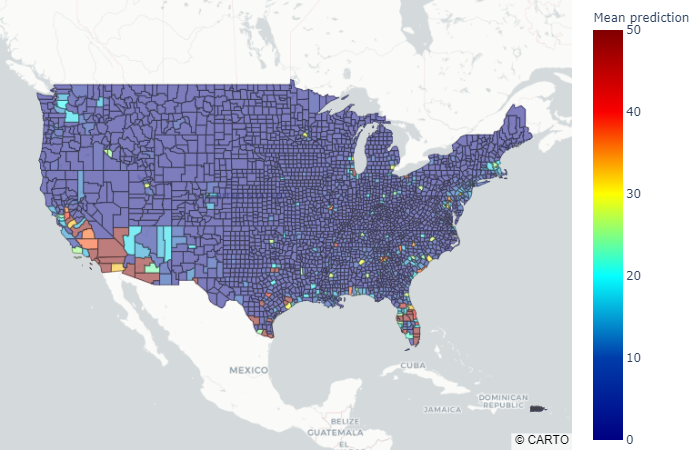}
 
 \includegraphics[width=.6\textwidth]{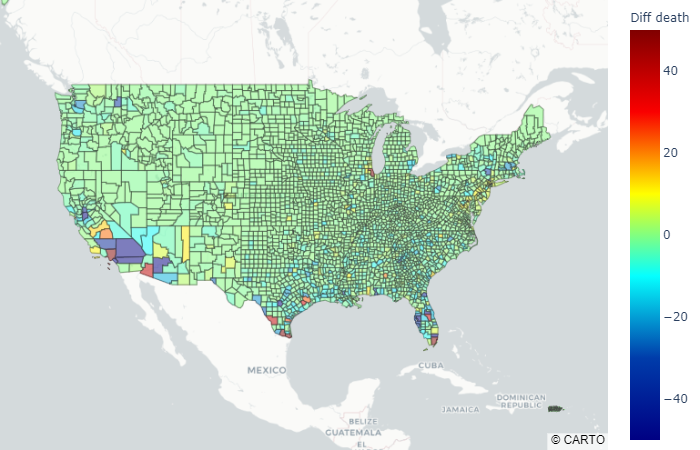}
   \caption{\textbf{County-wise prediction by condLSTM-Q.} County-wise absolute death counts (Top) and the differences with the ground truth (Bottom), summed from August 6 to August 19, 2020. This interval was roughly the peak of the ``second wave" of deaths in the United States. Note that the top and the bottom figures have different color scales.}
   \label{fig:ErrorMap}
\end{figure}

Visualization of the nationwide, county-based prediction from August 6 to August 19, 2020, is shown in Fig \ref{fig:ErrorMap}. This interval approximately covered periods of the COVID-19 ``second wave'' with the highest number of daily death attributed to COVID-19 in the United States, which impacted the Southwest and Southeast regions the most. With county-level resolution, condLSTM-Q demonstrated its ability to pick up inhomogeneous hot spots within specific states, whereas other coarser geographical models could only observe an average trend of the whole state. In addition, at this resolution, one can observe state-state interactions of spread in the state border. For instance, quite a few counties in Arizona and Nevada had trends more similar to their neighboring California hot spots than to other counties within the states.

\subsection{Comparison with other models}
To evaluate the performance of condLSTM-Q, we compared its predictions with the IHME model \cite{IHME}. As condLSTM-Q provides county-wise predictions while the IHME model generates state-wise predictions, we aggregated predictions of condLSTM-Q over each county into corresponding states and measured root mean square error (RMSE) of the state-wise forecasts on two-week intervals. We also averaged across the 9 quantile predictions from condLSTM-Q to estimate the mean of the distribution, in order to allow direct comparison against single value predictions from IHME. Note also that due to the format of the IHME model available, predictions from the two models are matched on a bimonthly basis.

\begin{figure}[h]
    \centering
    \begin{tabular}{cccc}
    \hline
    \multirow{2}{*}{Dates} & \multicolumn{3}{c}{14-day prediction RMSE}\\ \cline{2-4}
    & condLSTM-Q & \hspace{2em}IHME\hspace{2em} & \hspace{2em}control\hspace{2em} \\ \hline
    2020-05-04 & 54.814 & \textbf{38.189} & 67.881 \\ \hdashline
    2020-05-19 & \textbf{18.396} & 20.812 & 38.046 \\ \hdashline
    2020-06-03 & \textbf{13.709} & 14.152 & 27.152 \\ \hdashline
    2020-06-24 & \textbf{74.154} & 74.295 & 77.597 \\ \hdashline
    2020-07-04 & 16.291 & \textbf{14.790} & 28.945 \\ \hdashline
    2020-07-18 & 48.690 & \textbf{47.098} & 64.760 \\ \hdashline
    2020-08-06 & \textbf{32.927} & 33.749 & 55.354 \\ \hdashline
    2020-08-21 & \textbf{14.276} & 21.912 & 37.628 \\ \hdashline
    2020-09-02 & \textbf{16.325} & 18.915 & 34.328 \\ \hdashline
    2020-09-18 & \textbf{13.123} & 14.421 & 28.974 \\ \hline
    \end{tabular}
    \includegraphics[width=0.60\textwidth]{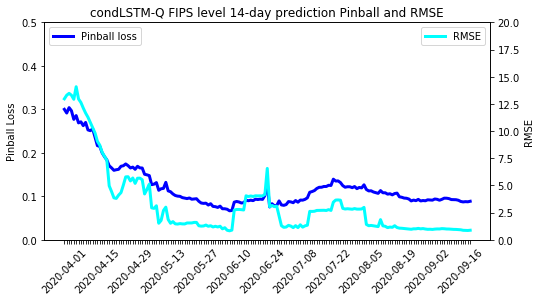}
    \caption{\textbf{Performance analysis.} State-level prediction (Top). State-wise, two-week prediction RMSE of condLSTM-Q and IHME model. We matched the starting dates of our predictions to the dates when the IHME model was updated, where each model would have access to the training data up to the day before the onset of the two-week prediction. We also tried placing zeros in all the predictions and calculated RMSE as a control. Both IHME and condLSTM-Q showed a similar pattern with the control, indicating that there is an uncaptured variance for both models. County(FIPS)-level performance measured by pinball loss and RMSE (Bottom). The error for both metrics showed a steady decline since April with several jumps from June to August, which is from the abnormality in New York Times’ data due to their bulk adjustments following the changes in tallying criteria.}
    \label{fig:error}
\end{figure}

As shown in Fig \ref{fig:error}, the state-wise predictions of condLSTM-Q were consistently comparable to the IHME model. The condLSTM-Q initially had higher RMSE in early May 2020, but showed better performance over most intervals since then after training data became ample. Despite the fact that condLSTM-Q model was trained using county-level data with the pinball loss function, it demonstrated a robust performance on a different geographical scale under a different metric (i.e. RMSE). We also measured the RMSE by setting zeros as control(i.e. placing zeros on the full 14-days interval across all 50 states in the United States). Our results revealed that both IHME and our condLSTM-Q model presented a similar pattern with the control, indicating that there could be an uncaptured variance in both models. We also noticed that there was an abrupt peak of the epidemic curve (approximately 20,000 deaths) on June 24, 2020, when New York Times made bulk adjustments in its data due to changes in tallying criteria.

\subsection{Usefulness of conditional layer for categorical data}

With a classical LSTM network which does not have a conditional layer for initializing the hidden states based on categorical data as in condLSTM-Q, the categorical data should be stacked to the same dimension of the time series data to be fed into the model. As mentioned in the Introduction, this may lead to suboptimal performance of the model because it introduces constant values in time series features which actually is not sequential data. To test this, we compared the prediction accuracy of two models: the model with the aforementioned stacking method, which we refer to as the  "pseudo-categorical" LSTM model, and condLSTM-Q. The pseudo-categorical LSTM model had a pinball loss of $0.115$ during the prediction period of May 22 - June 4, 2020, and $0.0994$ during June 5 - June 18, 2020. The pinball loss of condLSTM-Q, on the other hand, was $0.0959$ and $0.0744$, respectively, over the same prediction periods.

\begin{figure}[h]
    
    
    
    \centering
    \includegraphics[width=1\textwidth]{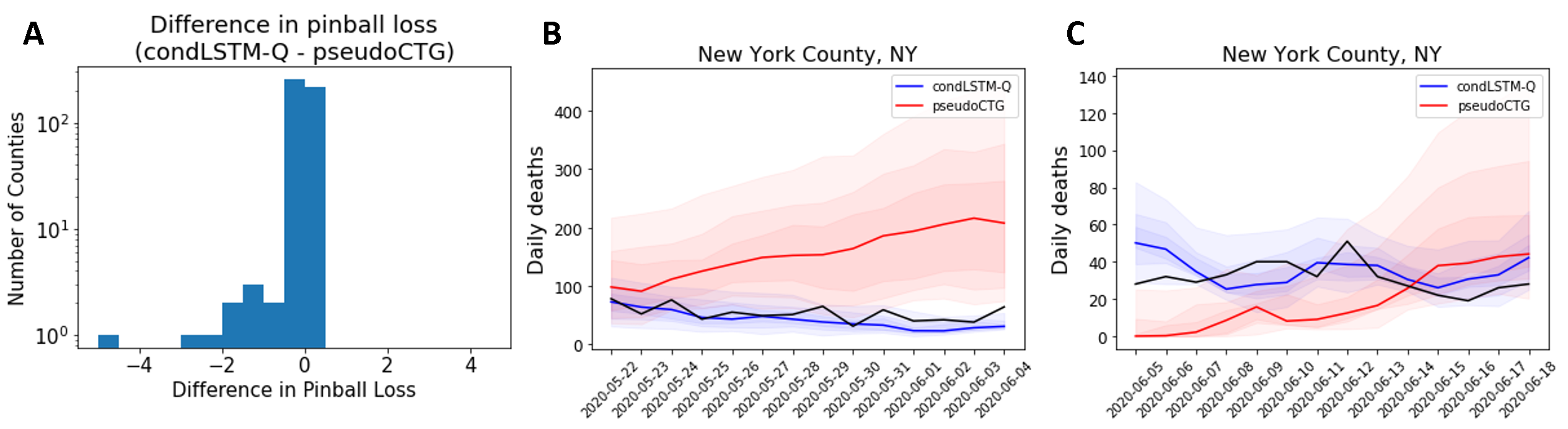}
    \caption{\textbf{Effectiveness of conditional architecture}. (A) A histogram of differences in pinball loss between condLSTM-Q and the pseudo-categorical model, in counties with a high number of total death (>50). The loss from the pseudo-categorical model was subtracted from condLSTM-Q's loss, so negative values mean condLSTM-Q has lower pinball loss. (B, C) Representative case study of New York, in two different time frames. The condLSTM-Q not only captures the trend better, but also has a tighter distribution when compared to the pseudo-categorical model trained on the same training data with categorical data stacked to match the time series.}
    \label{fig:condVSpseudoCTG}
\end{figure}

To investigate whether such a decrease in pinball loss was observed in every county, we looked at the difference in pinball loss between condLSTM-Q and pseudo-categorical LSTM model in each county. Fig \ref{fig:condVSpseudoCTG} A shows a distribution of such differences in counties with a high number of deaths ($>50$). The difference was obtained by subtracting the pinball loss of a pseudo-categorical model from condLSTM-Q’s loss, so its bias to the left means condLSTM-Q's loss is significantly lower than the pseudo-categorical model ($p<0.0005$, one-sided Wilcoxon signed-rank test)

The difference between these two models is well-demonstrated in Fig \ref{fig:condVSpseudoCTG} B and C, which shows example predictions of May 22 - June 4 and June 5 - June 18, 2020, in New York - the county with the largest difference in pinball loss between the models. When the condLSTM-Q was already capturing a down-turned trend, the pseudo-categorical LSTM was still predicting an upward trend. We also found that the condLSTM-Q had a much tighter spread of prediction, allowing planning based on the predictions much more possible. 


\subsection{Explainability of the model}


\begin{figure}[h]
    \centering
    \includegraphics[width=0.95\textwidth]{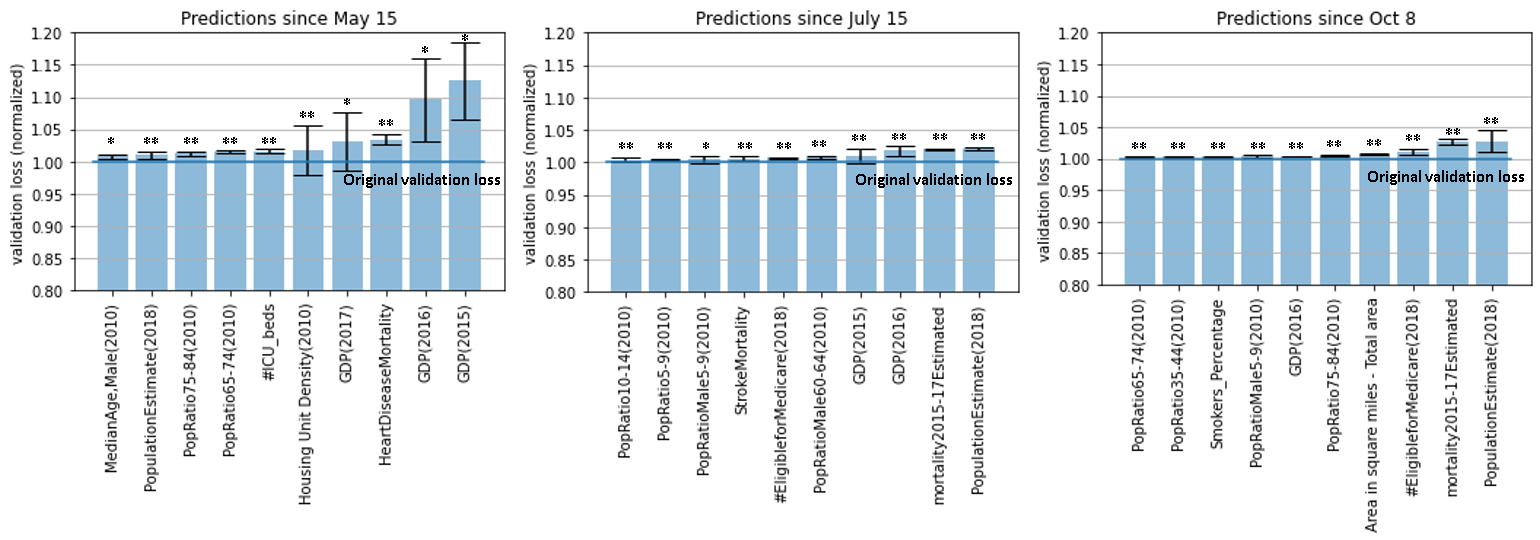}
    \caption{\textbf{Importance of each categorical feature}. The permutation feature importance of the 10 most important categorical features in three prediction windows (May 15-28, July 15-28, and October 8-21, 2020) is shown. The categorical features were more important during the earlier phase rather than in the later phase. The list of important features also changed over time: GDP and heart disease mortality for the earlier phase; population, mortality, and the number of eligible people for medicare for the later phase. Asterisks indicate statistical significance based on one-sample sign test to original validation loss (*p<0.05, **p<0.01)}
    \label{fig:shuffle_CTG}
\end{figure}

\begin{figure}[h]
    \centering
    \includegraphics[width=1\textwidth]{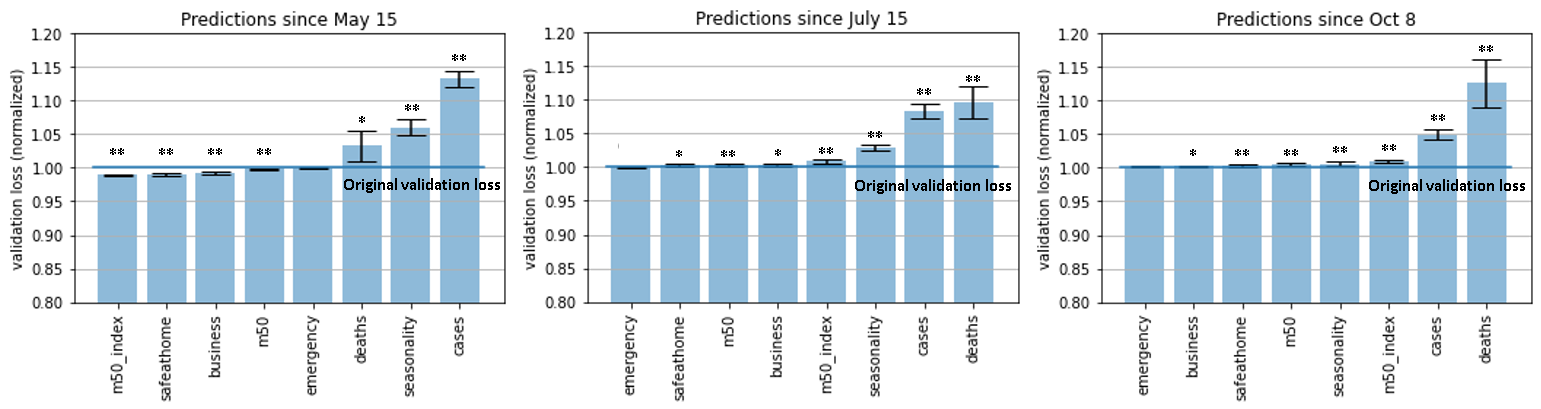}
    \caption{\textbf{Importance of each time series feature}. The permutation feature importance of the time series features in three prediction windows (May 15-28, July 15-28, and October 8-21, 2020) is shown. The number of cases, deaths, and flu seasonality are the three most important time series features in the model. The importance of seasonality, however, went down nearly to zero in the later phase. Asterisks indicate statistical significance from one-sample sign test to original validation loss (*p<0.05, **p<0.01)}
    \label{fig:shuffle_TS}
\end{figure}

To quantify the relative importance of individual features and see which has the most information for the prediction, permutation feature importance was measured for each time series and categorical feature. As suggested in previous studies \cite{shuffleTest_randomForest, shuffleTest_general}, we measured the feature importance based on how the model's validation loss has increased after permuting each feature in a validation set. Increased validation loss after shuffling a feature shows that the model is relying on the feature for the prediction, whereas an unchanged validation loss means the model is largely ignoring the feature during prediction.

For categorical data, each feature was shuffled across counties, as it does not have a temporal component. Time series features, on the other hand, were shuffled across counties first, and then permuted across time. As each county can have drastically different values in time series features, shuffling across counties and time separately ensures more realistic input data for LSTM, compared to the features shuffled across counties and time altogether. Such permutation was done 10 times for each feature, and the resulting 10 input data sets were given to the trained model to generate predictions. The validation loss was calculated from each prediction and compared to the original validation loss to see how the shuffled features affected the model's performance. As the importance of each feature can change over time, permutation feature importance was measured at three different time frames (May 15-28, July 15-28, and October 8-21, 2020) with the same procedure.

The result is shown in Fig \ref{fig:shuffle_CTG} and Fig \ref{fig:shuffle_TS}. For categorical data (Fig \ref{fig:shuffle_CTG}), its importance to the prediction was generally higher during the earlier phase, when there were insufficient time series data for the model to be trained on. The top three features with the greatest importance during the earlier phase in May were GDP in 2015, GDP in 2016, and heart disease mortality. GDP was also one of the most important features in the later phase in July, but it was not as important as the population estimate (2018) and estimated mortality (2015-2017). These two features were the most important categorical features in October as well, followed by the number of eligible people for Medicare (2018).

In contrast, the ranking of the importance of the time series features (see Fig \ref{fig:shuffle_TS}) was relatively stable compared to categorical features. The number of deaths and cases were the two most important features in all three time frames. In the beginning, the number of cases was more important, but was surpassed by the number of deaths later on. Seasonality also had high importance, especially during the earlier phase in May and July 2020 when there were fewer data for the model to be trained on.





\section{Discussion}

\subsection{Contribution and Novelty}

This is the first use of conditional RNN in the prediction of COVID-19 trend that we know of. The model has the merit of providing fine-scale, county-level predictions of daily deaths with a two-week window. The performance of the model was robust and comparable to, if not slightly better than, the well-known IHME predictions. Given that the application of this model type is very new in the field, and that we are far from testing all possible architectures with different input features, one could foresee that the model's performance in terms of stability, precision, and length of the forecast window could still be enhanced with further optimization. We also see that this type of model could be generalized to trend forecasting of other infectious diseases in the future and is not limited to COVID-19 itself.

County-level spatial resolution is a feature not available in other famous and publicly available forecasts such as IHME, DELPHI, and Los Alamos National Laboratory \cite{DELPHI, LANL} which provide state-wise predictions. This provides unique opportunities for investigating trends within the states, and interactions between counties along state borders. In addition, through the analysis of the importance of the categorical data which will be discussed below, one could learn which features are risk factors that affect the death trend and provide handles for officials to ameliorate the risks.

\subsection{Effect of categorical features on predictions}

\begin{figure}[h]
    \centering
    \includegraphics[width=0.95\textwidth]{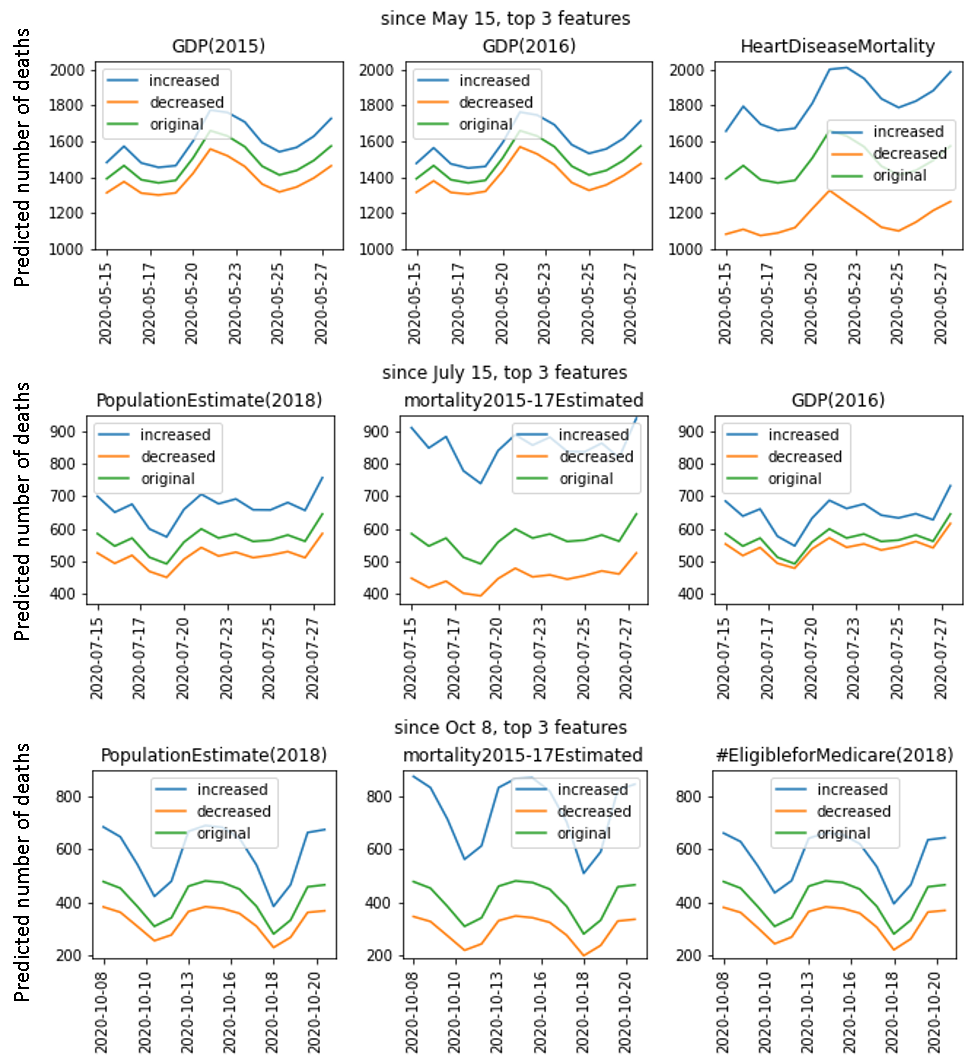}
    \caption{\textbf{Changes in nationwide prediction after modifying categorical features}. For each model making predictions over 2 weeks from May 15th (top), July 15th (middle), and October 8th, 2020 (bottom), new predictions after increasing and decreasing a categorical feature are shown in blue and orange, along with the original prediction in green. Each altered feature has been selected based on permutation feature importance from each of the models, and features were either increased or decreased by 3 standard deviations. The values shown in this figure are the predicted numbers of deaths in the United States.} 
    \label{fig:modelSensitivity}
\end{figure}

One advantage of using conditional LSTM is that it provides insight into the effect of each categorical and time series feature on the predictions. Fig \ref{fig:modelSensitivity} shows the total number of deaths predicted by each model after altering categorical features with the highest permutation feature importance. As mentioned in the Result, GDP in 2015 and 2016 and heart disease mortality were the most important features in the beginning, but later on, the dominant factors shifted to the population estimate, estimated mortality, and the number of eligible people for Medicare; an increase in any of these features led to an increase in predicted counts and vice versa.

Such a pattern in the earlier phase shows the possible vulnerability of people with heart disease to COVID-19, which is consistent with previous studies reporting elevated risk of poorer outcomes with COVID-19 in people with congenital heart disease \cite{CongenitalHeartDisease} and coronary heart disease \cite{CoronaryHeartDisease, CoronaryHeartDisease2, CoronaryHeartDisease3}. The positive correlation between GDP and the predicted number of deaths in the earlier phase is also consistent with a study claiming a positive correlation between GDP and the number of confirmed cases of COVID-19 in China \cite{GDPandCOVID} - nevertheless, the GDP estimates the value of goods and services produced by each county \cite{GDP}, which inevitably correlates with its population and other variables. Therefore, it would require further investigation to remove the dependencies of each factor.

In the later phase, on the other hand, the important features have shifted to population estimates, mortality rate, and the number of eligible people for Medicare. This may indicate that the number of deaths due to COVID-19 in the later phase is more related to the overall healthcare resources and general health condition of the population in each county.

\subsection{Additional Benefits}
The LSTM-based structure is scalable and flexible. We were limited to a few available data sources during the initiative and have not expanded actively to other data sources that possibly provide additional information for predictions, as testing the robustness of this architecture was the original goal of this pilot study. Once there are additional data sources, the condLSTM-Q could adapt and be retrained to extract information from the new data source without adjusting the architecture. For instance, if one hypothesizes that the temperature affects the death trend, the model can be easily retrained and tested with an additional time series feature of temperature added to the input data stream. One could also foresee that, when the vaccine is eventually distributed, the immunization rate in regions could be one additional important feature.


\section{Material Availability}\label{sec:materials}
Code and Model for our study are available in:
\newline
\url{https://github.com/cjackal/COVID-SKTW}

Demographics and Health Resource dataset from the Yu Group at UC Berkeley:
\newline
\url{https://raw.githubusercontent.com/Yu-Group/covid19-severity-prediction/data/county\_data\_abridged.csv}

COVID-19 mortality dataset from New York Times:
\newline
\url{https://raw.githubusercontent.com/nytimes/covid-19-data/master/us-counties.csv}

Local area GDP dataset from U.S. Bureau of Economic Analysis (BEA):
\newline
\url{https://www.bea.gov/news/2019/local-area-gross-domestic-product-2018}

Land area and Population density dataset from 2010 Census:
\newline
\url{https://factfinder.census.gov/faces/tableservices/jsf/pages/productview.xhtml?src=bkmk}

Mobility dataset from DescartesLabs:
\newline
\url{https://raw.githubusercontent.com/descarteslabs/DL-COVID-19/master/DL-us-mobility-daterow.csv}

COVID-19 related policies dataset from HHS:
\newline
\url{https://healthdata.gov/dataset/covid-19-state-and-county-policy-orders}

\section{Acknowledgement}
The authors thank Prof Yaser Abu-Mostafa, and the Teaching Assistants of CS156 in Caltech for organizing the COVID19 prediction initiative and for providing the data pipeline for parsing data sources. We thank Isaac Yen-Hao Chu M.D. for reading the manuscript. Y.L.Ni was supported by Taipei Veterans General Hospital-National Yang-Ming University Excellent Physician Scientists Cultivation Program, No.103-Y-A-003.

\bibliographystyle{unsrt}  
\bibliography{main}






\end{document}